\documentclass[10pt,twocolumn,letterpaper]{article}

\usepackage{wacv}
\usepackage{times}
\usepackage{epsfig}
\usepackage{graphicx}
\usepackage{amsmath}
\usepackage{amssymb}
\usepackage{booktabs}

\def\wacvPaperID{783} 

\wacvalgorithmstrack   

\wacvfinalcopy 

\ifwacvfinal
\usepackage[breaklinks=true,bookmarks=false]{hyperref}
\else
\usepackage[pagebackref=true,breaklinks=true,colorlinks,bookmarks=false]{hyperref}
\fi

\begin{document}

\title{Salient Temporal Encoding for Dynamic Scene Graph Generation}

\author{Zhihao Zhu\\
Carnegie Mellon University\\
{\tt\small 704242527zzh@gmail.com}}

\maketitle
\thispagestyle{empty}

\begin{abstract}
   Representing a dynamic scene using a structured spatial-temporal scene graph is a novel and particularly challenging task. 
   To tackle this task, it is crucial to learn the temporal interactions between objects in addition to their spatial relations. Due to the lack of explicitly annotated temporal relations in current benchmark datasets, most of the existing spatial-temporal scene graph generation methods build dense and abstract temporal connections among all objects across frames. However, not all temporal connections are encoding meaningful temporal dynamics. We propose a novel spatial-temporal scene graph generation method that selectively builds temporal connections only between temporal-relevant objects pairs and represents the temporal relations as explicit edges in the scene graph. The resulting sparse and explicit temporal representation allows us to improve upon strong scene graph generation baselines by up to $4.4\%$ in Scene Graph Detection. In addition, we show that our approach can be leveraged to improve downstream vision tasks. Particularly, applying our approach to action recognition, shows $0.6\%$ gain in mAP in comparison to the \mbox{state-of-the-art}.
\end{abstract}

\vspace{-0.5cm}
\section{Introduction}
\label{sec:intro}
A primary goal of computer vision is to have a detailed understanding of a visual scene comparable to what humans perceive. Recently, scene graph generation has emerged as a novel task in computer vision. In addition to correctly locating and recognizing objects in a static scene, it requires a perceptual algorithm to infer the semantic interactions between them. Scene graph generation has first been proposed in \cite{ImageRetrieval@Justin}, and great progress has been made towards improving its performance \cite{KnowledgeEmbeddedRoutingNetwork, lin2020gps, FullyConvolutionalSceneGraphGeneration, IterativeMessagePassing@Danfei, GrapgRCNN@Jianwei}. 

\begin{figure}[t]
\centering
\includegraphics[width=0.5\textwidth]{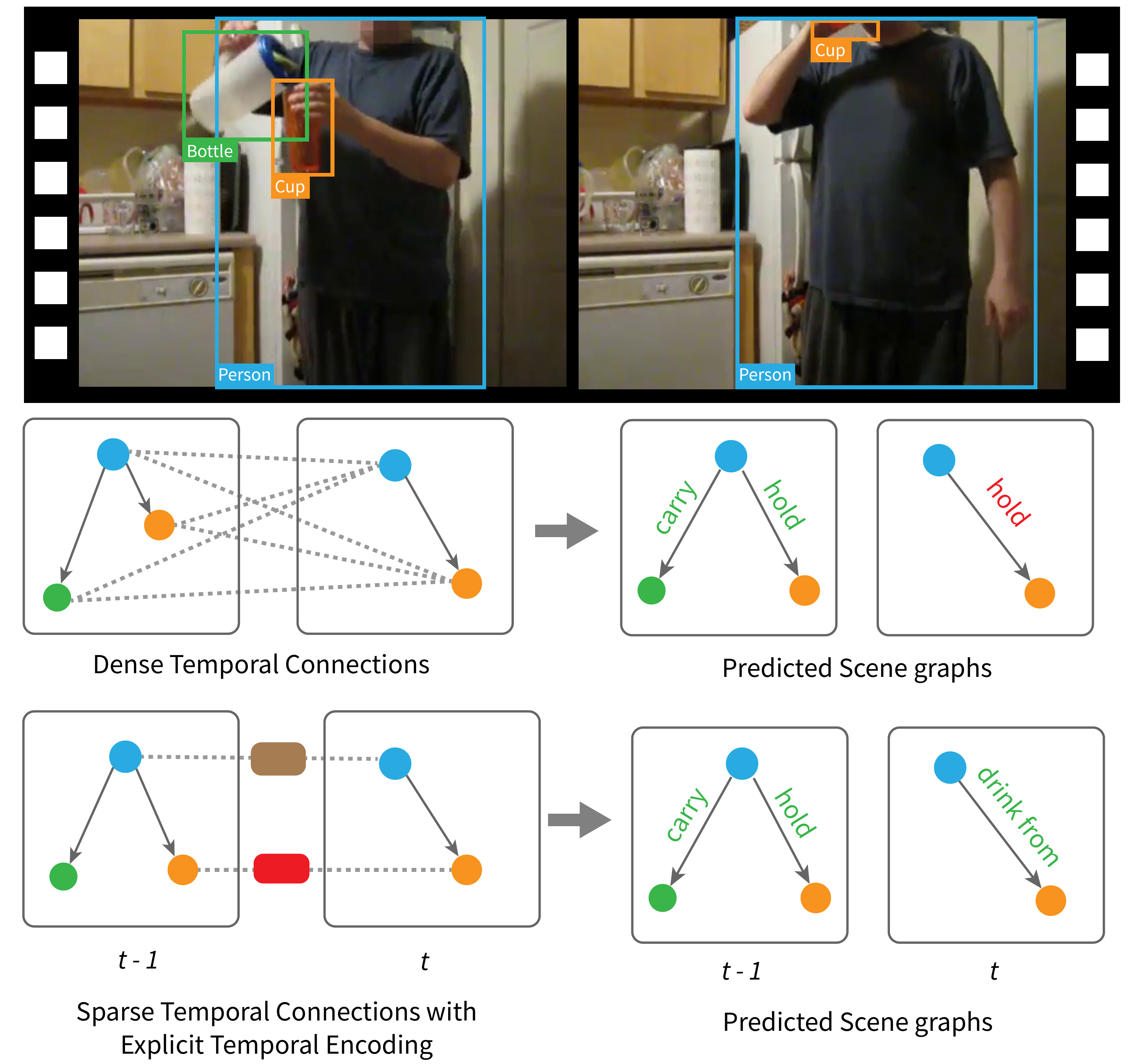}
\vspace{-0.1cm}
\caption{Top row shows a person pouring water into a cup and then drinking the water from the cup. Second and third row compare state-of-the-art dense temporal connections and our sparse and explicit temporal connections for dynamic scene graph generation. Circular nodes (blue, green and orange) represent the detected objects, and rectangular nodes (brown and red) represent our encoded temporal relations.}
\label{tab:teaser_image}
\vspace{-0.5cm}
\end{figure}

While spatial scene graphs are an effective representation of static scenes, they lack the ability to model temporal dynamics of visual elements across time. Temporal context plays a crucial role in action and activity understanding. By observing a single frame of a human-object interaction, it is often difficult, even for a person, to accurately identify the action. This indicates the importance for scene understanding algorithms to analyze not a single frame but a sequence of frames, in order to correctly understand interactions.

\begin{figure*}[t]
\includegraphics[width=1\textwidth]{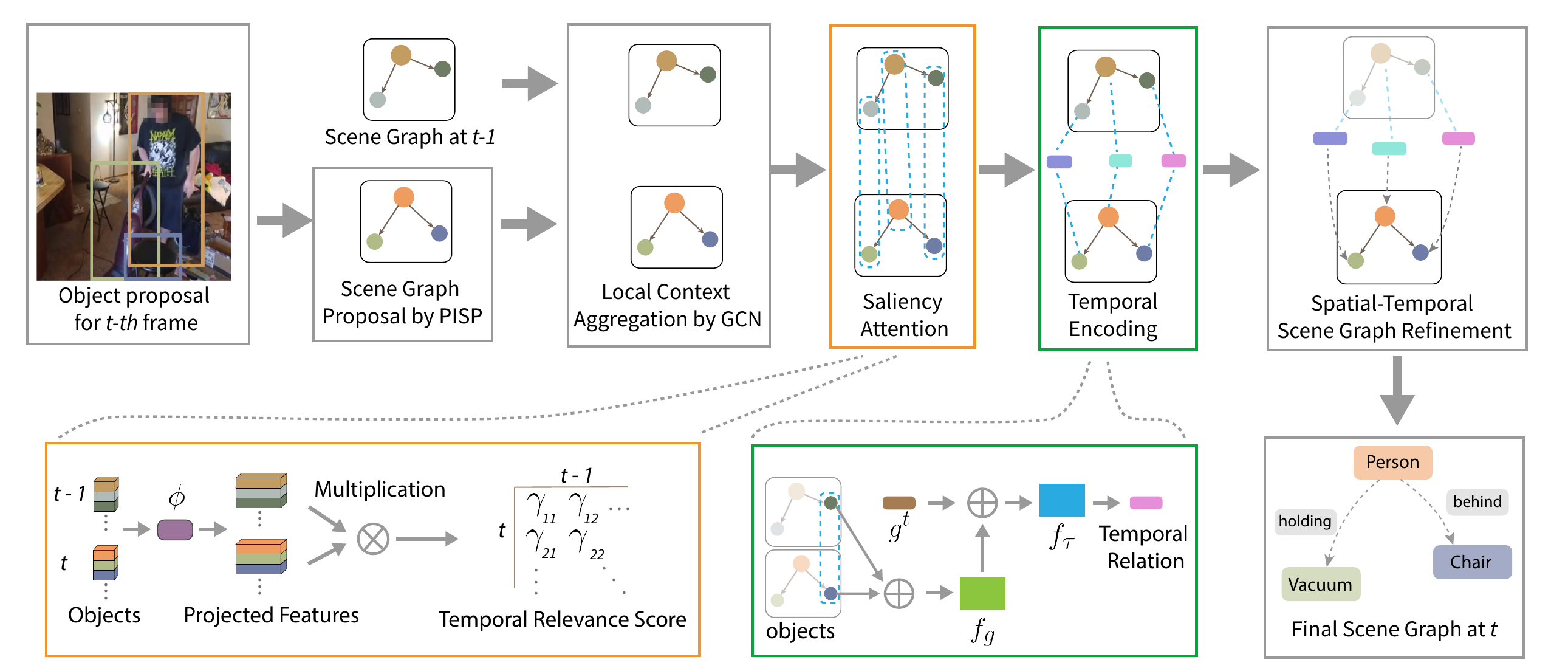}
\caption{Overview of Salient Temporal Relation Encoder (STRE). From left to right, frame $t$ is input into PISP \cite{herzig2018mapping} to obtain an initial spatial scene graph proposal. To aggregate spatial context for each node, the spatial scene graph prediction from the previous frame $t-1$ are input into a Graph Convolutional Network. The updated node features are passed into the Saliency Attention module to calculate temporal relevance score. The pairs with the highest temporal relevance scores are selected for encoding the temporal relation via the Temporal Relation Encoder. The spatial context and encoded temporal relations are summed up and passed into the spatial-temporal scene graph refinement module to obtain the final graph.}
\label{Fig:ModelArchitecture}
\vspace{-0.5cm}
\end{figure*}

To mitigate the limits of static scene graphs for representing a dynamic scene, Jingwei et al. \cite{ji2019action} proposed spatial-temporal scene graphs. 
Individual spatial scene graphs computed for each frame are stacked together to form the spatial-temporal scene graph. 
Therefore, this approach doesn't explicitly model temporal dynamics, which are important for understanding the semantics of a dynamic scene. For instance, in Fig. \ref{tab:teaser_image}, the cup's location changes from under the bottle to near the person's face. This temporal change is helpful in inferring the semantics that the person is pouring water from the bottle into the cup in frame $t-1$, and then drinks water from the cup in frame $t$.

We propose Salient Temporal Relationship Encoder (STRE) to leverage these temporal dynamics by explicitly modelling them as additional features in the spatial-temporal scene graph. Temporal features encode dynamic changes of objects and context across frames. To the best of our knowledge, our approach is the first to model spatial-temporal context as feature vectors for scene graph generation. These features can be used for downstream tasks and we show how they can improve action recognition.

While learning temporal context is important, we posit that a limited set of temporal connections across frames contains meaningful information - temporal connections should be established sparsely, between objects that contribute to the same event of the scene.
In the example given in Fig. \ref{tab:teaser_image}, our proposed STRE (bottom row) does not connect the bottle in frame $t-1$ to every object in frame $t$ because the bottle doesn't contribute to the main event in the frame $t$ that the person is drinking water from the cup.
This is in contrast to previous work \cite{arnab2021unified, cong2021spatial, Wang2018VideosAS} that builds dense temporal connections (upper row) among all the objects across frames. Experiments in Section \ref{exp:temporal_scene_graph} confirm that by selectively building only the most important temporal connections, STRE yields significant gains in performance and efficiency compared to recent state of the art.

Details of Salient Temporal Relation Encoder (STRE) are outlined in Fig. \ref{Fig:ModelArchitecture}. The contributions of our work can be summarized as follows: (i) we propose the use of sparse spatial-temporal graphs to only build the most important temporal connections and show state of the art performance on scene graph generation, while being more efficient, (ii) a method of  explicitly encoding spatial-temporal context into features that can be used in downstream tasks, (iii) a novel way to incorporate the spatial-temporal scene graph into action recognition to better leverage temporal dynamics and improve action recognition.

\section{Related Work}
\paragraph{Spatial Scene Graphs}
Recently, structured scene representations have had much attention in the computer vision community. Static scene graphs introduced in \cite{ImageRetrieval@Justin} have been applied to solve different vision tasks. Scene graphs represent images by explicitly modeling objects and their relationships to create a graph where objects are represented as nodes and relations as edges. This structure can both be used for scene understanding \cite{gao2020drg, herzig2018mapping, lin2020gps, Tang2019LearningTC, GrapgRCNN@Jianwei,  zellers2018neural} and for downstream tasks \cite{ImageCaptioningwithSceneGraph, ImageGeneration@Justin, santoro2017simple}. An increase in uses for scene graph contextual reasoning has also been proliferated by new scene graph datasets \cite{krishna2016visual, openimages2020, liang2019vrrvg}. These datasets allow myriad of approaches towards improving scene graph generation. Some approaches use pairwise methods \cite{ herzig2018mapping, li2017scene,  santoro2017simple, Tang2019LearningTC, zhang2019graphical, zhu2018automatic} while others use recurrence \cite{IterativeMessagePassing@Danfei, zellers2018neural}, GNNs \cite{GrapgRCNN@Jianwei}, or transformers \cite{geng2021dynamic, lin2020gps}. Despite great progress made by these methods for scene graph generation, their applications on videos are limited, as they are designed for static images.
\vspace{-0.5cm}
\paragraph{Spatial-Temporal Scene Graphs}
Spatial-temporal graphs have been proposed to improve upon spatial scene graphs by considering the interactions between objects over time. 
Recent datasets \cite{ji2019action, Rai2021HomeAG, shang2017video} in this area have mostly focused on human-object interactions. 
Many approaches have been proposed to solve the spatial-temporal scene graph generation task by applying different temporal context aggregation strategies. For example, some encode temporal context using recurrent \cite{xietowards} or pooling \cite{geng2021dynamic} methods. Others use graph convolutions \cite{arnab2021unified, geng2021dynamic, Wang2018VideosAS} and attention \cite{cong2021spatial, geng2021dynamic, Ji_2021_ICCV, teng2021target}. However, most of these models explicitly separate the spatial and temporal interaction to sequential modules, limiting their expressibility. 
More recent methods achieved state-of-the-art performance on the Action Genome dataset \cite{ji2019action} by using the spatial-temporal transformer \cite{cong2021spatial}, or the message passing graph neural network \cite{arnab2021unified}. However, these methods need to build dense temporal connections, which is inefficient and obscures important temporal dynamics. In contrast, our proposed STRE builds sparse temporal connections only between relevant objects. Experiments demonstrate that this leads to better temporal contextual understanding and more accurate spatial-temporal scene graph generation. 
Recently Li et al. \cite{li2022dynamic} avoids building dense temporal connections by matching objects in different frames only if they are spatially-close. However, this approach has limits in cases when objects are moving fast, or when there are many objects located closely in one region. In contrast, 
our model selectively builds meaningful sparse temporal connections, by considering whether the established temporal connections contribute to the main event in the scene. Experiments in Section \ref{exp:temporal_scene_graph} show our model's advantage over \cite{li2022dynamic}'s hand-crafted criteria for building temporal connections. 
\vspace{-0.5cm}
\paragraph{Video Action Recognition}
Video action recognition has evolved rapidly in recent years. Common approaches focus primarily on utilizing RGB or flow-based features \cite{Carreira2017QuoVA, Feichtenhofer2019SlowFastNF, Hussein2019TimeceptionFC, Wu2019LongTermFB}. Recently, Baradel et al. \cite{Baradel2018ObjectLV} has found benefit of using spatial interaction of detected objects in a frame as additional features for action recognition. Spatial-temporal interactions between objects have also been shown \cite{arnab2021unified,Materzynska2020SomethingElseCA, Actor-Centric, Wang2018VideosAS, ou2022recognition, ryoo2021tokenlearner} to further improve action recognition performance. In particular, \cite{ji2019action} suggests using structured spatial-temporal scene graph as the feature bank \cite{Wu2019LongTermFB} for action recognition. However, they stack multiple static scene graphs to construct feature banks, which lack the ability to model temporal dynamics. While also using \cite{Wu2019LongTermFB} as backbone like in \cite{ji2019action}, our model explicitly encodes a temporal scene graph feature bank, which helps the action recognition model leverage temporal context more easily.

\section{Salient Temporal Relation Encoder}

The goal of the Salient Temporal Relation Encoder (STRE) is to encode sparse spatial-temporal context as a feature and use this context to inform scene graph prediction. A common approach in literature \cite{herzig2019spatio, ji2019action, Materzynska2020SomethingElseCA, xietowards} to generate a spatial-temporal scene graph is to first generate a spatial scene graph for each individual frame and then stack them along the time axis.

A static spatial scene graph $SSG = (\mathcal{V}, \mathcal{E})$ is a structured representation of the content of an image ${I}$, where $\mathcal{V}$ is a set of nodes that correspond to objects in the image, and $\mathcal{E}$ is a set of edges representing relationships between nodes. 
Time adds dependencies between nodes in different frames $\boldsymbol{I} = \{I^1, I^2, ...\}$. A spatial-temporal scene graph is often defined as: $STSG = (\mathcal{V}^t, \mathcal{E}^t)$, introducing an additional temporal index $t$. To be able to explicitly leverage the temporal context, we propose a new spatial-temporal scene graph representation $STSG = (\mathcal{V}^t, \mathcal{E}^t, \mathcal{T}^{t-1, t})$, where $\mathcal{T}^{t-1, t} = \{\tau_{11}^{t-1, t}, \tau_{12}^{t-1, t}, ...\}$ is the temporal context, and each element $\tau_{ij}^{t-1, t}$ 
represents the temporal dependencies between the $i$-th object in frame $I^{t-1}$ and $j$-th object in frame $I^t$.

Given a video, our approach generates spatial-temporal scene graph in three-stages: 
(i) spatial scene graph proposal using a well-tuned spatial scene graph generation network \cite{herzig2018mapping} (Section \ref{SpatialSceneGraph}), 
(ii) temporal scene graph generation (Section \ref{TemporalSceneGraph}), and 
(iii) joint spatial-temporal scene graph refinement (Section \ref{STSGRefinement}).

For application on the action recognition task, a long term feature bank is constructed from the final spatial-temporal scene graphs. Details are discussed in Section \ref{STSGActionRecognition}. 
\subsection{Spatial Scene Graph}
\label{SpatialSceneGraph}

A spatial scene graph is required as an initial proposal to define the nodes and edges per frame. 
We choose the Permutation-Invariant Structured Prediction model (PISP) \cite{herzig2018mapping} as a spatial baseline and an initial prediction of the spatial scene graph. We find this architecture simple and straightforward to integrate temporal features. Additionally, PISP can be tuned to perform similar to state-of-the-art for the spatial-only prediction on the Action Genome dataset.

For a given frame $I^t$, a Faster R-CNN object detector \cite{Ren2015FasterRT} is used to extract a set of object proposals, where each object proposal $i$ consists of a box location \mbox{$b_i = [ x_i, y_i, w_i, h_i ]$}, object label distributions $p_i$ over $G$ object classes, and $C$-dimensional ROI-Aligned visual features $\mu_i \in \mathbb{R}^{C}.$ 
Each node $\nu_i \in \mathcal{V}$ is represented as the concatenation of box location, label distribution and visual features, namely  $\nu_i =(b_i \oplus p_i \oplus \mu_i) \in \mathbb{R}^{M}$, where $\oplus$ is the concatenation operation, and $M=4+G+C$. To represent the initial guess of the relations between $N$ number of objects, a relation distribution matrix $R \in \mathbb{R}^{N \times N \times H}$ is constructed for each object pair, where $H$ is the number of relationship classes.



\subsection{Temporal Scene Graph}
\label{TemporalSceneGraph}

To build only meaningful temporal connections, we propose a two-stage temporal scene graph generation method. The first stage,  Saliency Attention, sparsely connects the top K node pairs from two spatial scene graphs based on contextual similarity. The second stage applies a Temporal Relation Encoder to generate spatial-temporal features between node pairs.

\vspace{-0.2cm}
\subsubsection{Saliency Attention}
\label{SaliencyAttention}
In order to build compact and effective temporal scene graphs, the most temporally-relevant node pairs need to be identified. We claim that, meaningful temporal dynamics are between either the same objects across frames, or different objects if they contribute to the same event of the scene. To achieve this, we propose a novel Saliency Attention module to calculate temporal relevance scores between objects across frames, and build temporal connections between node pairs that have highest scores. The calculation of temporal relevance score takes into account the similarity of object features as well as their context. The latter of which is important for determining whether two objects contribute to the same event.


To build spatial-temporal context across objects, a modified Graph Convolutional Network (GCN) \cite{GCN} is applied to the spatial scene graph proposal. The modified GCN aggregates both neighboring node information $\nu_j$ and relation information $\varepsilon_{i,j}$ to obtain an updated node representation $\hat{\nu}_i \in \mathbb{R}^{M}$. More specifically, one propagation layer can be represented as: 
\begin{equation}
\hat{\nu}_i = \sigma(\nu_i + \sum_{j \in \mathcal{N}(i)}\theta_{ij}(W_{\nu}\nu_j+W_{\varepsilon}\varepsilon_{ij}) ),
\label{eq:important}
\end{equation}
where $\sigma$ is a non-linear activation function, $\mathcal{N}(i)$ is the first order neighborhood of node $i$, $\theta_{ij} \in [0, 1]$ represents the connectivity between node $i$ and $j$, where $0$ means the two nodes are not connected. $W_{\nu}: \mathbb{R}^{M} \to \mathbb{R}^{M}$ and $W_{\varepsilon}: \mathbb{R}^{L} \to \mathbb{R}^{M}$ are two learned linear layers that transform the nodes and edges respectively into the same feature dimension.

We use a temporal relevance score to determine if objects should be connected (they involve the same spatial-temporal interaction). To do this, we sort pairs of objects by node feature similarity and select the $K$ highest-ranked temporal relevance scores. $K$ is a parameter that controls the sparsity of the temporal scene graph, and a smaller $K$ results in sparser temporal scene graph. A common method to compute similarity is to concatenate two node features and use a Multi-Layer Perceptron (MLP). This requires calculation for each pair of nodes, which results in quadratic $O(N^{t-1} \times N^t)$ time complexity. To reduce time complexity, a kernel function $f_{\gamma}(.)$ is applied instead. Given two GCN-updated node features $\hat{\nu}_j^t$ and $\hat{\nu}_k^{t-1}$ for the frame $I^t$ and $I^{t-1}$ respectively, the temporal relevance score $\gamma_{jk}$ is calculated as:
\begin{equation}
\gamma_{jk} = f_{\gamma}(\hat{\nu}_j^{t}, \hat{\nu}_{k}^{t-1}) = \left \langle \phi(\hat{\nu}_{j}^{ t}), \phi(\hat{\nu}_{k}^{t-1}) \right \rangle ,
\label{eq:important}
\end{equation}
where $\phi: \mathbb{R}^{M} \to \mathbb{R}^{D}$ is a learned projection function which projects nodes features into $D$-dimensional space. 



\vspace{-0.2cm}
\subsubsection{Temporal Relation Encoding}
\label{TemporalEncoding}
We propose a Temporal Relation Encoder to explicitly model the temporal context as features and not attention weights like \cite{arnab2021unified, cong2021spatial}. To achieve this, the frame context $g^t$ is defined as the average of the GCN-updated node features: $g^t = \frac{1}{N^t}\sum_{i=1}^{N^t} \hat{\nu}_{i}^{t}$, which is leveraged to guide the temporal relation encoding.

More specifically, for every node pairs $j, k$ that's connected by Saliency Attention (i.e. $\eta_{jk}=1$), their node features $\hat{\nu}_{j}^{t}$ and $\hat{\nu}_{k}^{t-1}$ are concatenated, and projected into a common space to form the global context feature $g^t$. Together with $g^t$, they are fed into a feature encoder to encode the temporal relations $\tau_{jk}^{t, t-1}$.
We adopted a MLP as the temporal relation encoder. More specifically, temporal relations $\tau_{jk}^{t, t-1}$ can be represented as:
\begin{equation}
\tau_{jk}^{t, t-1} = f_{\tau}(g^t \oplus f_{g}(\hat{\nu}_{j}^{t} \oplus \hat{\nu}_{k}^{t-1})),
\label{eq:important}
\end{equation}
where $\oplus$ is the concatenation operation, and $f_{\tau}: \mathbb{R}^{2M} \to \mathbb{R}^{M}$ and $f_{g}: \mathbb{R}^{2M} \to \mathbb{R}^{M}$ are both two-layer MLPs. 
A detailed comparison of different encoders (e.g Transformer) is provided in Section 2 of the supplemental material.


\subsection{Spatial-Temporal Scene Graph Refinement}
\label{STSGRefinement}
Finally, to improve the per-frame scene graph prediction from the initial proposal we use the spatial-temporal features. For each node $j$, its temporal relations are aggregated with a weighted sum over all its valid temporal connections, with using temporal relevance score as the weights. Aggregated temporal relation $\zeta_j^t$ can be represented as: 
\vspace{-0.1cm}
\begin{equation}
\zeta_j^t = \frac{1}{\sum_{k=1}^{N^{t-1}}\gamma_{jk}\cdot\eta_{jk}}\sum_{k=1}^{N^{t-1}}\gamma_{jk}\cdot\tau_{jk}^{t, t-1}\cdot \eta_{jk},
\label{eq:important}
\end{equation}

The spatial and temporal context are leveraged to form the final node feature representation $\vartheta_j^t$. To obtain $\vartheta_j^t$, aggregated temporal relations $\zeta_j^t$ and GCN-updated node features $\hat{\nu}_j^t$ are summed up:
\begin{equation}
\vartheta_j^t = \hat{\nu}_j^t + \zeta_j^t,
\label{eq:finalFeature}
\end{equation}

To estimate the scene graph relations, the updated subject node feature $\vartheta_i^t$ and object node feature $\vartheta_j^t$ are concatenated, and passed into a fully-connected layer $f_{r}$ followed by a sigmoid activation function $\sigma$, resulting in the relation class distribution between subject $i$ and object $j$:
\begin{equation}
p_{ij}^t = \sigma(f_{r}(\vartheta_i^{t} \oplus \vartheta_j^t)),
\label{eq:important}
\end{equation}


\subsection{Spatial-Temporal Scene Graph for Action Recognition}
\label{STSGActionRecognition}
An additional benefit of sparse and explicit temporal relations is that they can be easily leveraged for downstream action recognition. To validate this, similar to \cite{ji2019action}, we use \cite{Wu2019LongTermFB} as our baseline action recognition model. In \cite{ji2019action}, the authors constructed a Scene Graph Feature Bank (SGFB) by stacking a sequence of spatial scene graphs, excluding any explicit temporal graph  dynamics.

To integrate explicit temporal dynamics, we propose a new Spatial-Temporal Scene Graph Feature Bank (ST-SGFB): $Z_{ST-SGFB} = [z^1, z^2, ..., z^T]$ which leverages both spatial and temporal relations. More specifically, each component $z^t$ is a $|K \times M|$ matrix, given there are $K$ number of object classes, and each object has a $M$-dimensional feature representation. If an object of class $i$ is detected in the frame $I^t$, the $i$-th row of $z^t$ will be updated with $\vartheta_i^t$, which is the aggregation of the spatial context and temporal context as shown in Eq. \ref{eq:finalFeature}. If there is no detected object for the $i$-th class, then the $i$-th row in the $z^t$ will be set to zero. We follow the same approach as \cite{ji2019action} to use ST-SGFB as the long-term feature bank together with the 3D CNN features for classifying the actions.

\section{Experiments}
In Section \ref{exp:temporal_scene_graph} the Salient Temporal Relation Encoder (STRE) is evaluated for the spatial-temporal scene graph generation task. 
In Section \ref{exp:action_recognition} the Spatial-Temporal Scene Graph Feature Bank 
is evaluated for the action recognition task. 
\subsection{Spatial-Temporal Scene Graph Generation}
\label{exp:temporal_scene_graph}
\subsubsection{Dataset, Training, and Evaluation}{
\paragraph{Dataset} Recently two video-based scene graph datasets with frame-level scene graph annotations have been released (i) the Action Genome dataset \cite{ji2019action} which is based upon well established Charades dataset \cite{sigurdsson2016hollywood} and (ii) the Home Action Genome dataset \cite{Rai2021HomeAG}, a large-scale multi-view video dataset of daily activities at home. We evaluate scene graph generation on the Action Genome dataset \cite{ji2019action}, as the Home Action Genome dataset \cite{Rai2021HomeAG} has not been used yet in literature for evaluation of scene graph generation performance, while Action Genome dataset allows for comparison with the widest selection of these approaches. 

\vspace{-0.3cm}


\begin{table}[ht]
  \fontsize{7.9}{7.2}\selectfont
  \centering
  \begin{tabular}{cp{0.5cm}p{0.6cm}p{0.5cm}p{0.6cm}p{0.5cm}p{0.6cm}cccc}
  \midrule

                         & \multicolumn{2}{c}{PredCls} & \multicolumn{2}{c}{SGCls} & \multicolumn{2}{c}{SGDet} \\
    \cmidrule(lr){2-3}
    \cmidrule(lr){4-5}
    \cmidrule(lr){6-7}
                         & R@10 & R@20 & R@10 & R@20 & R@10 & R@20 \\
  \midrule\midrule

  \multicolumn{1}{c|}{VRD \cite{lu2016visual}} & 51.7 & 54.7 & 32.4 & 33.3 & 19.2 & 24.5
  \\

  \multicolumn{1}{c|}{Motif Freq \cite{zellers2018neural}}  & 62.4 & 65.1 & 40.8 &41.9 &  23.7 & 31.4
  \\

  \multicolumn{1}{c|}{MSDN \cite{li2017scene}} & 65.5 & 68.5 & 43.9 & 45.1 & 24.1 & 32.4
  \\

  \multicolumn{1}{c|}{VCTREE \cite{Tang2019LearningTC}} & 66.0 & 69.3 & 44.1 & 45.3 & 24.4 & 32.6
  \\

  \multicolumn{1}{c|}{RelDN \cite{zhang2019graphical}} & 66.3 & 69.5 & 44.3 & 45.4 & 24.5 & 32.8
  \\

  \multicolumn{1}{c|}{GPS-Net \cite{lin2020gps} } & 66.8 & 69.9 & 45.3 & 46.5 & 24.7 & 33.1
  \\

  \midrule
  
  \multicolumn{1}{c|}{STTran \cite{cong2021spatial}} & 68.6 & 71.8 & 46.4 & 47.5 & 25.2 & 34.1
  \\
  
  \multicolumn{1}{c|}{APT \cite{li2022dynamic}} & \textbf{69.4} & \textbf{73.8} & 47.2 & \textbf{48.9} & 26.3 & 36.1
  \\

  \multicolumn{1}{c|}{STRE (ours)} & \textbf{69.4} &  71.4 & \textbf{47.4} & 48.5 & \textbf{29.0} & \textbf{38.4}
  \\

  \bottomrule\bottomrule
  \vspace{0.1cm}
  \end{tabular}
    \caption{\textit{With-constraint} comparison with state-of-the-art spatial (upper) and spatial-temporal (lower) scene graph generation methods on the Action Genome dataset \cite{ji2019action}. All other approaches use the same detector and following the protocol suggested in \cite{cong2021spatial}.} 
  \vspace{-0.2cm}
  \label{tab:full_results}
\end{table}

\begin{table}[ht]
  \fontsize{6.9}{2.2}\selectfont
  \centering
  \begin{tabular}{cp{0.5cm}p{0.6cm}p{0.5cm}p{0.6cm}p{0.5cm}p{0.6cm}cccc}
  \midrule

                         & \multicolumn{2}{c}{PredCls} & \multicolumn{2}{c}{SGCls} & \multicolumn{2}{c}{SGDet} \\
    \cmidrule(lr){2-3}
    \cmidrule(lr){4-5}
    \cmidrule(lr){6-7}
                         & mR@20 & mR@50 & mR@20 & mR@50 & mR@20 & mR@50 \\
  \midrule\midrule

  \multicolumn{1}{c|}{Freq Prior \cite{ZoomNet}} & 55.1 & 63.6 & 34.3 & 36.6 & 24.8 & 34.0
  \\

  \multicolumn{1}{c|}{G-RCNN \cite{GrapgRCNN@Jianwei}}  & 56.3 & 61.3 & 36.1 & 38.2 &  27.7 & 34.9
  \\

  \multicolumn{1}{c|}{RelDN \cite{zhang2019graphical}} & 59.8 & 63.4 & 39.9 & 41.9 & 30.3 & 39.5
  \\

  \multicolumn{1}{c|}{TRACE \cite{TRACE}} & 61.8 & 65.3 & 41.1 & 43.2 & 30.8 & 40.1
  \\
   
  \midrule
  
  \multicolumn{1}{c|}{STRE(Ours)} & \textbf{62.7} & \textbf{67.5} & \textbf{41.9} & \textbf{44.2} & \textbf{32.4} & \textbf{42.7}
  \\
   
  \bottomrule\bottomrule
  \vspace{0.1cm}
  \end{tabular}
    \caption{Mean recall(\%) of various models on AG. The number of triplets per frame is set to a limit of 50 and top 6 predictions for each pair are kept when evaluating.} 
  \vspace{-0.2cm}
  \label{tab:mrecall}
\end{table}

\paragraph{Training} To train the model, a sample is constructed from a series of frames. 
Since the ratio of relations to no relations across objects is low, the relation loss is divided by the number of relations to ensure equal weight for relation and no-relation predictions. In addition the weight of no-relation predictions is decreased by a factor of 5.

For classification of the relations a binary cross-entropy loss is applied. In addition STRE can refine the object class label predicted by the detector by utilizing an entity node classification loss in the SGCls and SGDet modes. Additional training details (including hyper parameters) can be found in the supplemental.
\vspace{-0.3cm}
\paragraph{Evaluation} Following \cite{ji2019action}, we adopt three common recall-based evaluation protocols for image-based scene graph generation: (i) \textit{Predicate Classification} (PredCls), (ii) \textit{Scene Graph Classification} (SGCls), and 
(iii) \textit{Scene Graph Detection} (SGDet). 
Cong et al. \cite{cong2021spatial} recently provided a unified experimental protocol using the same object detector and evaluation protocol for different baseline methods to ensure comparability with previous work \cite{ji2019action,Newell2017PixelsTG, zellers2018neural, cong2021spatial}. We are following the same evaluation protocol, and use their provided detector model checkpoint to obtain the same object detection results. 

In addition, to better account for the imbalanced distribution of relations, we also evaluate our model using mean Recall(mR), $\text{mAP}_{\text{rel}}$ and $\text{wmAP}_{\text{rel}}$, following \cite{TRACE}.

}
\vspace{-0.3cm}
\subsubsection{Quantitative Results and Comparison}{

\begin{figure*}[!th]
\includegraphics[width=1\textwidth]{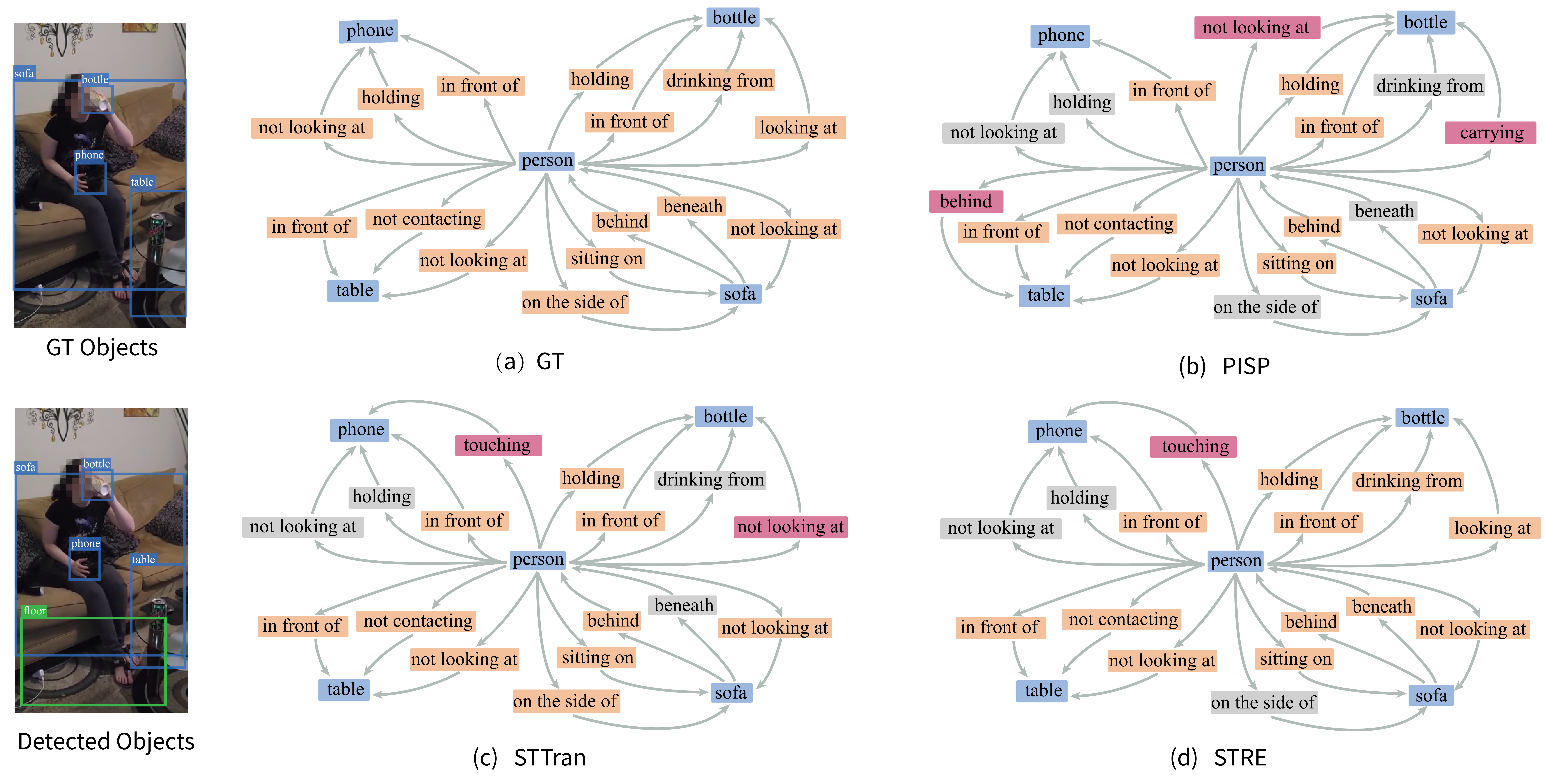}
\vspace{-0.4cm}
\caption{Qualitative comparison of scene graphs generated by: (b) PISP\cite{herzig2018mapping}, (c) STTran\cite{cong2021spatial} and (d) our STRE model. Green box in the bottom-left image indicates a detected floor which does not exist in the ground-truth. For better visualization, only the top 12 relations from \textit{no constraint} results are plotted, excluding the relations with the floor. The orange-colored relations are correct predictions, the red-colored relations are incorrect predictions that do not exist in the ground-truth, and the gray-colored relations are missing predictions.}
\label{Fig:SGQualitativeResults}
\vspace{-0.3cm}
\end{figure*}

Table \ref{tab:full_results} compares STRE to several state-of-the-art scene graph generation methods on the Action Genome dataset \cite{ji2019action}. Results using \textit{no constraint} setting are available in the supplemental. STRE is trained using $K=1$ as suggested by detailed performance analysis of this parameter in Section \ref{exp:sparse_vs_dense}. 

\begin{table}[!ht]
  \fontsize{7.9}{2.2}\selectfont
  \centering
  \begin{tabular}{cp{0.4cm}p{0.6cm}p{0.4cm}p{0.6cm}p{0.4cm}p{0.9cm}cccc}
  \midrule

                         & \multicolumn{2}{c}{PredCls} & \multicolumn{2}{c}{SGCls} & \multicolumn{2}{c}{SGDet} \\
    \cmidrule(lr){2-3}
    \cmidrule(lr){4-5}
    \cmidrule(lr){6-7}
                         & mAP\_r & wmAP\_r & mAP\_r & wmAP\_r & mAP\_r & wmAP\_r \\
  \midrule\midrule

  \multicolumn{1}{c|}{Freq Prior \cite{ZoomNet}} & 33.1 & 65.9 & 14.2 & 22.6 & 9.4 & 15.5
  \\

  \multicolumn{1}{c|}{G-RCNN \cite{GrapgRCNN@Jianwei}}  & 41.2 & 70.8 & 17.6 & 22.5 & 11.7 & 15.5
  \\

  \multicolumn{1}{c|}{RelDN \cite{zhang2019graphical}} & 50.1 & 72.2 & 20.1 & 23.8 & 12.9 & 15.9
  \\

  \multicolumn{1}{c|}{TRACE \cite{TRACE}} & \textbf{53.2} & 75.2 & 20.7 & 24.6 & 13.4 & 16.5
  \\
   
  \midrule
  
  \multicolumn{1}{c|}{STRE(Ours)} & 53.1 & \textbf{80.1} & \textbf{21.0} & \textbf{26.7} & \textbf{13.9} & \textbf{17.9}
  \\
   
  \bottomrule\bottomrule
  \vspace{0.1cm}
  \end{tabular}
  \caption{ $mAP_{rel}$ and $wmAP_{rel}$ (\%) of various models on AG. The number of triplets per frame is set to a limit of 50 and top 6 predictions for each pair are kept when evaluating.} 
  \vspace{-0.2cm}
  \label{tab:mAP}
\end{table}

Experimental results confirm that by sparsely building temporal connections and explicitly encoding temporal dynamics, STRE is able to outperform other methods across almost all the metrics. A significant gain is observed compared to STTran \cite{cong2021spatial} by $4.3\%$ in SGDet-R@20. 
In addition STRE outperforms the recent Anticipatory Pre-Training (APT) method \cite{li2022dynamic} by $2.3\%$ in SGDet-R@20 
, despite ATP using additional unlabeled data.

While the same object detector is used for STRE as STTran \cite{cong2021spatial}, STRE consistently outperforms STTran\cite{cong2021spatial} in the \mbox{SGCls} and SGDet metrics. This shows the strength of our STRE model to overcome the limitations of the object detector (e.g. miss-classifications of bounding boxes, false positive detections) when generating scene graphs, by better leveraging temporal dynamics across frames. For the \mbox{PredCls} metric, smaller performance gain of $0.8\%$ at Recall@10 is observed. 
When compared to SGCls and SGDet metrics, PredCls assumes perfect object detection predictions, which makes temporal association between frames much easier compared to object detector outputs. 

In addition, Table \ref{tab:mrecall} compares different models using mean recall, and
Table \ref{tab:mAP} compares models using $\text{mAP}_{\text{rel}}$ and $\text{wmAP}_{\text{rel}}$. These are more balanced metrics for datasets with imbalanced class distribution, thus better revealing model's capability. Results show that STRE improves over current state-of-the-art approaches in most metrics. Specifically, STRE outperforms TRACE\cite{TRACE} by 2.6\% and 1.4\% on SGDet mR@50 and SGDet $\text{wmAP}_{\text{r}}$.

\vspace{-0.3cm}
\subsubsection{Qualitative Results and Comparison}{Figure \ref{Fig:SGQualitativeResults} shows a qualitative comparison of dynamic scene graphs generated by STRE, the baseline PISP \cite{herzig2018mapping}, and the state-of-the-art STTran \cite{cong2021spatial}. We primarily compare against STTran\cite{cong2021spatial} as it is a strong baseline, which is also the most similar in training methodology, and has publicly available code and detectors to generate qualitative results and ensure a fair comparison. Results indicate that STRE improves capturing the important temporal dependencies among objects, and generates more accurate scene graphs. For example, STRE predicts fewer incorrect relations (colored in red), while also having fewer missed relations (colored in gray). Specifically this example shows that our STRE approach is able to correctly infer the main event in the scene: \textit{person - drinking from - bottle}, while both PISP and STTran\cite{cong2021spatial} failed to capture that relation. Despite the overall better quality of the generated scene graph, our model also has some limitations. For example, between person and phone, \textit{not looking at} and \textit{holding} haven't been captured. This might be due to the lack of clear visual clue, as the phone looks blended into the person's clothes, which makes it hard to correctly infer the correct relations.
}
\begin{figure}[t]
\centering \includegraphics[width=0.45\textwidth]{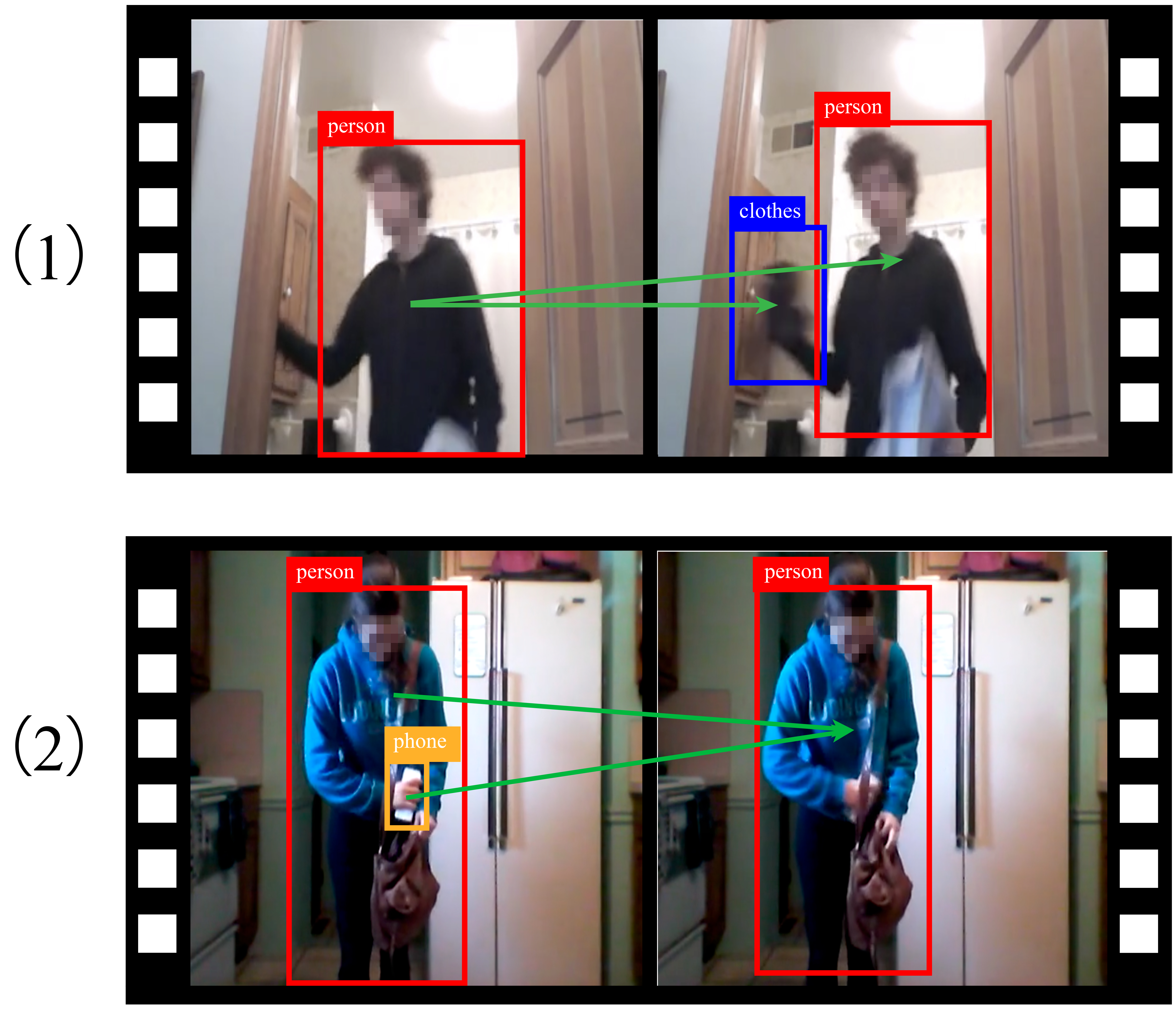}
\caption{Two examples showing temporal connections between different objects across frames predicted by STRE. Green arrows represent temporal connections.}
\label{Fig:Temporal_Connection_Example}
\end{figure}

\renewcommand{\arraystretch}{1}
\begin{table}[tb]
    \fontsize{8.9}{8.7}\selectfont
    \centering
    \setlength\tabcolsep{3pt}
    \begin{tabular}{l | c | c | c }
         & PredCls-R@20 & SGCls-R@20 & SGDet-R@20\\ 
        \midrule
        PISP \cite{herzig2018mapping}  & 70.9 & 48 & 33.1 \\
        \midrule
        STRE[K=1]  & 71.4 & 48.5 & 38.4 \\
        STRE[K=2]  & 71.4 & 48.4 & 38.1 \\
        STRE[K=3]  & 71.3 & 48.4 & 38 \\
        \midrule
        DTRE & 71.0 & 48.2 & 37.4\\
    \end{tabular}
    \vspace{0.1cm}
     \caption{Performance comparison of PISP \cite{herzig2018mapping}, STRE and DTRE for scene graph prediction. All metrics are \textit{with constraint}.} 
    \vspace{-0.5cm}
    \label{tab:scene_graph_ablation}
\end{table}
\vspace{-0.2cm}
\subsubsection{Sparse vs Dense Temporal Connections}
\label{exp:sparse_vs_dense} To evaluate the advantage of the sparse temporal connections, we compare: (i) PISP \cite{herzig2018mapping}, our  baseline spatial scene graph generation model without modeling any temporal relations, (ii) different level of sparsity of STRE by varying the numbers of temporal connections $K$, and (iii) Dense Temporal Relation Encoder (DTRE). To enable DTRE, Saliency Attention is modified to allow temporal connections between all node pairs across two frames. Based on our experiment, on average DTRE has 8 temporal connections per object.

Experiments in Table \ref{tab:scene_graph_ablation} confirm that by leveraging temporal context, STRE and DTRE both improve performance over the PISP model across all metrics. Further, sparser STRE (smaller $K$) outperforms denser STRE (larger $K$). For instance, STRE[K=1] has a performance improvement over STRE[K=3] ranging from $0.1\%$ to $0.4\%$ across all metrics. Especially, compared with fully-connected temporal graph DTRE, STRE[K=1] has an improvement of $1\%$ in SGDet-R@20, which shows that meaningful temporal dependencies do not exist between all object pairs. This confirms that capturing the most relevant temporal relations is important for generating more accurate scene graphs.

To highlight the benefit on model efficiency when using sparse temporal connections, we compare the number of parameters and FLOPs to the \cite{cong2021spatial} in Table \ref{tab:efficiency}. Our model has significantly smaller number of parameters and FLOPs, and requires less computation than \cite{cong2021spatial}, while still achieving better performance both quantitatively and qualitatively.
\vspace{-0.2cm}
\subsubsection{Tracking vs Saliency Attention for Building Temporal Connections} 
As discussed in Section \ref{TemporalSceneGraph}, while tracking can only connect the same object across frames, Saliency Attention is able to also connect different objects as long as they contribute to the same event of the scene. This is possible because the scene graph is updated by a GCN before feeding into Saliency Attention, and thus node features contain local context, which helps to determine whether two nodes are under the same context(contributing to the same event).

For instance, in the upper example in Figure \ref{Fig:Temporal_Connection_Example}, STRE is able to build a temporal connection between the clothes in the second frame and the person in the first frame, as they both contribute to the event: ``the person fetches the clothes". Similarly, in the bottom example in Figure \ref{Fig:Temporal_Connection_Example}, STRE connects the phone in the first frame to the person in the second frame, as they are major objects in the event ``the person puts phone away". A tracking algorithm would fail to build such temporal connections for the clothes in the first example, and the phone in the second, due to occlusions.
\renewcommand{\arraystretch}{1}
\begin{table}[tb]
    \small
    \centering
    \begin{tabular}{l | c | c }
         & \#Parameters (M)  &  \#FLOPs (M)/frame\\ 
        \hline
        STTran \cite{cong2021spatial}  & 91.8 & 577  \\
        STRE[K=1]  & 21.7 & 120\\
    \end{tabular}
    \vspace{0.1cm}
    \caption{Efficiency comparison between our STRE[K=1] and STTran \cite{cong2021spatial}. The object detector network is not included. FLOPs/frame computation is estimated for three detected objects per frame.}
    \vspace{-.6cm}
\label{tab:efficiency}
\end{table}

Table \ref{tab:tracking_vs_STRE} shows the quantitative performance comparison between STRE using Saliency Attention (we set $K=1$) and STRE using a detection-based tracker to build temporal connections. 
The tracker connects each object in current frame  to an object in the following frame that has the same detection label, and is spatially closest.
Saliency attention based STRE outperforms the tracking-based STRE across all metrics and achieves $0.8\%$ improvement on \mbox{SGDet-R@20} compared to the tracking approach.
\renewcommand{\arraystretch}{.7}
\begin{table}
    \fontsize{8.9}{8.7}\selectfont
    \centering
    \setlength\tabcolsep{3pt}
    \begin{tabular}{l | c | c | c }
         & PredCls-R@20 & SGCls-R@20 & SGDet-R@20\\
        \midrule
        Tracking-Based & 70.7 & 48 & 37.8\\
        STRE  & 71.4 & 48.5 & 38.4 \\
    \end{tabular}
    \vspace{0.1cm}
    \caption{Performance comparison of Salient Temporal Relation Encoder (STRE) and replacing Saliency Attention with tracking (first row), for scene graph prediction. All metrics are \textit{with constraint}. } 
    \vspace{-0.2cm}
    \label{tab:tracking_vs_STRE}
\end{table}

Additional ablation studies on model architectures including Spatial Scene Graph Proposal and Temporal Relation Encoder can be found in the supplementary materials.
}


\setlength{\tabcolsep}{2pt}
\renewcommand{\arraystretch}{0.8}
\begin{table}[t]
    \small
    \centering
    \begin{tabular}{p{0.15\textwidth}p{0.15\textwidth}p{0.15\textwidth}}
    Method & Backbones & mAP \\ 
    \hline
    STAG \cite{herzig2019spatio} & R50-I3D &  37.2\\
    I3D + NL \cite{Carreira2017QuoVA, Wang2018NonlocalNN} & R101-I3D-NL & 37.5\\
    STRG \cite{Wang2018VideosAS} & R101-I3D-NL & 39.7\\
    SlowFast \cite{Feichtenhofer2019SlowFastNF} & R101 & 42.1\\
    LFB \cite{Wu2019LongTermFB} & R101-I3D-NL & 42.5\\
    ASF \cite{ZhangLM21} & R101-I3D-NL & 44.2 \\
    \hline
    SGFB \cite{ji2019action} & R101-I3D-NL & 44.3\\
    OR$^2$G \cite{ou2022recognition} & R101-I3D-NL & 44.9\\
    S-SGFB & R101-I3D-NL & 45.1\\
    ST-SGFB  & R101-I3D-NL & \textbf{45.5}\\
    
    \end{tabular}
    \vspace{0.1cm}
    \caption{Action Recognition on Charades dataset. All backbone models are pretrained on Kinetics-400 \cite{kay2017kinetics}.}
    \label{tab:action_recognition}
    \vspace{-0.3cm}
\end{table}

\subsection{Video Action Recognition}
\label{exp:action_recognition}
\subsubsection{Charades Dataset, Training, and Evaluation}{
Charades \cite{sigurdsson2016hollywood} is a commonly used, large-scale video action recognition dataset. It contains $9,848$ videos with $80\%$ of the videos in the training set. There are $157$ total action classes and $46$ object types.
For comparability, all experiments use the sampling procedure and per frame annotations described in \cite{ji2019action}. We pre-train with Kinetics-400 and cache scene graph predictions from STRE. To train for action classification, we follow the same protocol as \cite{ji2019action} by sampling a subset of frames and using the standard binary cross entropy loss. More details are added in supplemental.

}
\vspace{-0.3cm}
\subsubsection{Quantitative Results and Comparisons}
\begin{figure}[t]
\includegraphics[width=0.5\textwidth]{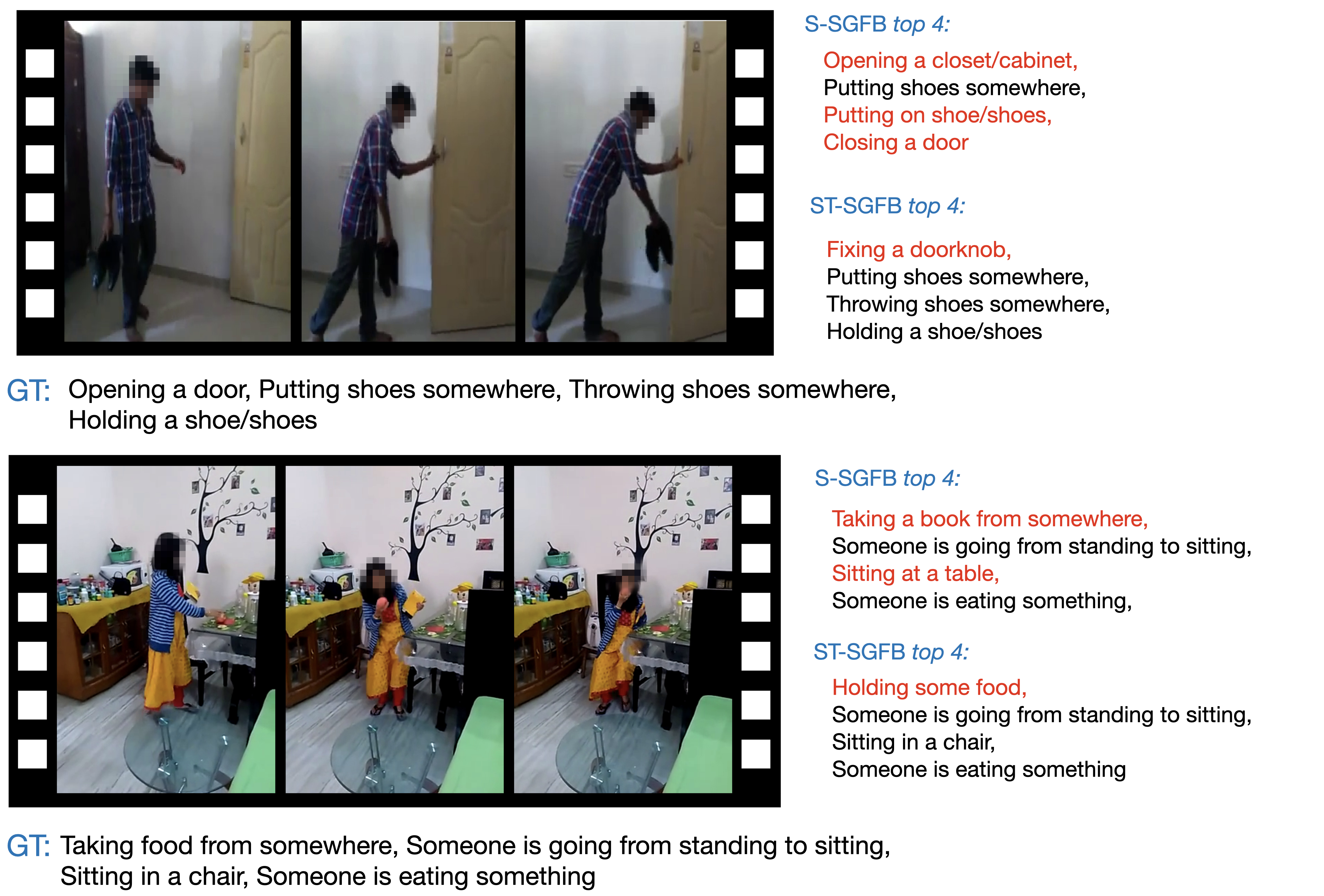}
\caption{Action Recognition qualitative results from two models: (1) S-SGFB which uses spatial scene graph as feature bank, (2) ST-SGFB which uses spatial-temporal scene graph as feature bank. Red means wrong predictions.}
\label{tab:Action_recognition_result1}
\vspace{-0.5cm}
\end{figure}

{
To verify the benefit of our STRE for downstream action recognition two experiments are conducted to evaluate the advantage of both the (i) spatial scene graphs and (ii) spatial-temporal scene graphs generated by our model. For (i), we follow \cite{ji2019action} to construct a Spatial Scene Graph Feature Bank by stacking a sequence of spatial scene graphs(i.e. without temporal edges). While some recent methods \cite{ryoo2021tokenlearner, MoViNets, assemblenet, assemblenetplusplus} perform better, we choose \cite{ji2019action} as our main baseline as it integrates scene graphs to improve action recognition and can be used in an online manner, unlike \cite{ryoo2021tokenlearner} which is offline. 

To avoid confusion with SGFB \cite{ji2019action} we denote our Spatial Scene Graph Feature Bank as S-SGFB. For (ii), we establish the Spatial-Temporal Scene Graph Feature Bank (ST-SGFB), as detailed in Section \ref{STSGActionRecognition}, by leveraging both spatial scene graphs as well as temporal scene graphs.

Table \ref{tab:action_recognition} shows the quantitative comparison to several state-of-the-art action recognition approaches. In particular, our direct comparison is SGFB \cite{ji2019action}, since we all apply scene graphs on the same baseline action recognition model: LFB \cite{Wu2019LongTermFB}. Experiments demonstrate a $2.6\%$ and $0.8\%$ improvement in mAP over LFB \cite{Wu2019LongTermFB} and SGFB \cite{ji2019action} respectively, by using our Spatial Scene Graph Feature Bank (S-SGFB). This highlights that by leveraging salient temporal context, the suggested STRE approach generates more accurate spatial scene graphs, which improves action recognition performance. Using ST-SGFB further increased the performance by $0.4\%$ compared to using our S-SGFB. This demonstrates that the explicitly encoded temporal relations in the scene graph feature bank are beneficial for the action recognition task.
\vspace{-0.6cm}
\subsubsection{Qualitative Results and Comparisons}
Figure \ref{tab:Action_recognition_result1} shows that ST-SGFB improves qualitatively compared to S-SGFB. In the left example, S-SGFB captures only one correct action label with respect to the shoes: \textit{putting shoes somewhere}. In comparison, ST-SGFB  correctly predicts all three actions, by leveraging the temporal dynamic change of the person-shoes relations from holding it to putting it away. In the right example, S-SGFB  wrongly predicts the person is \textit{taking a book from somewhere}. ST-SGFB overcomes this false prediction by recognizing that there is little temporal changes of the person's relation to the book: as the person is always holding the book. 



}
\vspace{-0.2cm}

\section{Conclusion}
We proposed a novel and efficient way for modeling sparse, explicit spatial-temporal interactions in dynamic scene graphs. Experiments demonstrated how Saliency Attention and Temporal Relation Encoding helped improve spatial-temporal scene graph predictions. In addition, we showed that using predicted spatial-temporal scene graphs can additionally improve action recognition performance. Future work could explore sparse temporal connections across more frames, or further improvement in model efficiency.

{\small
\bibliographystyle{ieee_fullname}
\bibliography{main}

\begin{thebibliography}{10}\itemsep=-1pt

\bibitem{arnab2021unified}
Anurag Arnab, Chen Sun, and Cordelia Schmid.
\newblock Unified graph structured models for video understanding.
\newblock In {\em ICCV}, 2021.

\bibitem{Baradel2018ObjectLV}
Fabien Baradel, Natalia Neverova, Christian Wolf, Julien Mille, and Greg Mori.
\newblock Object level visual reasoning in videos.
\newblock In {\em ECCV}, 2018.

\bibitem{Carreira2017QuoVA}
Jo{\~a}o Carreira and Andrew Zisserman.
\newblock Quo vadis, action recognition? a new model and the kinetics dataset.
\newblock In {\em CVPR}, pages 4724--4733, 2017.

\bibitem{KnowledgeEmbeddedRoutingNetwork}
Tianshui Chen, Weihao Yu, Riquan Chen, and Liang Lin.
\newblock Knowledge-embedded routing network for scene graph generation.
\newblock In {\em CVPR}, 2019.

\bibitem{cong2021spatial}
Yuren Cong, Wentong Liao, Hanno Ackermann, Bodo Rosenhahn, and Michael~Ying Yang.
\newblock Spatial-temporal transformer for dynamic scene graph generation.
\newblock In {\em ICCV}, 2021.

\bibitem{Feichtenhofer2019SlowFastNF}
Christoph Feichtenhofer, Haoqi Fan, Jitendra Malik, and Kaiming He.
\newblock Slowfast networks for video recognition.
\newblock In {\em ICCV}, pages 6201--6210, 2019.

\bibitem{gao2020drg}
Chen Gao, Jiarui Xu, Yuliang Zou, and Jia-Bin Huang.
\newblock Drg: Dual relation graph for human-object interaction detection.
\newblock In {\em ECCV}, 2020.

\bibitem{ImageCaptioningwithSceneGraph}
Lizhao Gao, Bo Wang, and Wenmin Wang.
\newblock Image captioning with scene-graph based semantic concepts.
\newblock In {\em {ICMLC}}, 2018.

\bibitem{geng2021dynamic}
Shijie Geng, Peng Gao, Moitreya Chatterjee, Chiori Hori, Jonathan~Le Roux, Yongfeng Zhang, Hongsheng Li, and Anoop Cherian.
\newblock Dynamic graph representation learning for video dialog via multi-modal shuffled transformers.
\newblock In {\em AAAI}, 2021.

\bibitem{herzig2019spatio}
Roei Herzig, Elad Levi, Huijuan Xu, Hang Gao, Eli Brosh, Xiaolong Wang, Amir Globerson, and Trevor Darrell.
\newblock Spatio-temporal action graph networks.
\newblock In {\em International Conference on Computer Vision Workshops}, 2019.

\bibitem{herzig2018mapping}
Roei Herzig, Moshiko Raboh, Gal Chechik, Jonathan Berant, and Amir Globerson.
\newblock Mapping images to scene graphs with permutation-invariant structured prediction.
\newblock In {\em NeurIPS}, 2018.

\bibitem{Hussein2019TimeceptionFC}
Noureldien Hussein, Efstratios Gavves, and Arnold W.~M. Smeulders.
\newblock Timeception for complex action recognition.
\newblock In {\em CVPR}, pages 254--263, 2019.

\bibitem{Ji_2021_ICCV}
Jingwei Ji, Rishi Desai, and Juan~Carlos Niebles.
\newblock Detecting human-object relationships in videos.
\newblock In {\em ICCV}, pages 8106--8116, October 2021.

\bibitem{ji2019action}
Jingwei Ji, Ranjay Krishna, Li Fei-Fei, and Juan~Carlos Niebles.
\newblock Action genome: Actions as compositions of spatio-temporal scene graphs.
\newblock In {\em CVPR}, pages 10236--10247, 2020.

\bibitem{ImageGeneration@Justin}
Justin Johnson, Agrim Gupta, and Li Fei{-}Fei.
\newblock Image generation from scene graphs.
\newblock In {\em CVPR}, 2018.

\bibitem{ImageRetrieval@Justin}
Justin Johnson, Ranjay Krishna, Michael Stark, Li{-}Jia Li, David~A. Shamma, Michael~S. Bernstein, and Li Fei{-}Fei.
\newblock Image retrieval using scene graphs.
\newblock In {\em CVPR}, 2015.

\bibitem{kay2017kinetics}
Will Kay, Joao Carreira, Karen Simonyan, Brian Zhang, Chloe Hillier, Sudheendra Vijayanarasimhan, Fabio Viola, Tim Green, Trevor Back, Paul Natsev, Mustafa Suleyman, and Andrew Zisserman.
\newblock The kinetics human action video dataset, 2017.

\bibitem{GCN}
Thomas~N. Kipf and Max Welling.
\newblock Semi-supervised classification with graph convolutional networks.
\newblock In {\em ICLR}, 2017.

\bibitem{MoViNets}
Dan Kondratyuk, Liangzhe Yuan, Yandong Li, Li Zhang, Mingxing Tan, Matthew Brown, and Boqing Gong.
\newblock Movinets: Mobile video networks for efficient video recognition.
\newblock In {\em {IEEE} Conference on Computer Vision and Pattern Recognition, {CVPR} 2021, virtual, June 19-25, 2021}. Computer Vision Foundation / {IEEE}, 2021.

\bibitem{krishna2016visual}
Ranjay Krishna, Yuke Zhu, Oliver Groth, Justin Johnson, Kenji Hata, Joshua Kravitz, Stephanie Chen, Yannis Kalantidis, Li-Jia Li, David~A. Shamma, Michael~S. Bernstein, and Fei-Fei Li.
\newblock Visual genome: Connecting language and vision using crowdsourced dense image annotations.
\newblock In {\em IJCV}, 2016.

\bibitem{openimages2020}
Alina Kuznetsova, Hassan Rom, Neil Alldrin, Jasper Uijlings, Ivan Krasin, Jordi Pont-Tuset, Shahab Kamali, Stefan Popov, Matteo Malloci, Alexander Kolesnikov, and et al.
\newblock The open images dataset v4.
\newblock In {\em IJCV}, 2020.

\bibitem{li2017scene}
Yikang Li, Wanli Ouyang, Bolei Zhou, Kun Wang, and Xiaogang Wang.
\newblock Scene graph generation from objects, phrases and region captions.
\newblock In {\em ICCV}, 2017.

\bibitem{li2022dynamic}
Yiming Li, Xiaoshan Yang, and Changsheng Xu.
\newblock Dynamic scene graph generation via anticipatory pre-training.
\newblock In {\em CVPR}, 2022.

\bibitem{liang2019vrrvg}
Yuanzhi Liang, Yalong Bai, Wei Zhang, Xueming Qian, Li Zhu, and Tao Mei.
\newblock {VrR-VG}: Refocusing visually-relevant relationships.
\newblock In {\em ICCV}, 2019.

\bibitem{lin2020gps}
Xin Lin, Changxing Ding, Jinquan Zeng, and Dacheng Tao.
\newblock {GPS-Net}: Graph property sensing network for scene graph generation.
\newblock In {\em CVPR}, 2020.

\bibitem{FullyConvolutionalSceneGraphGeneration}
Hengyue Liu, Ning Yan, Masood~S. Mortazavi, and Bir Bhanu.
\newblock Fully convolutional scene graph generation.
\newblock In {\em CVPR}, 2021.

\bibitem{lu2016visual}
Cewu Lu, Ranjay Krishna, Michael Bernstein, and Li Fei-Fei.
\newblock Visual relationship detection with language priors.
\newblock In {\em ECCV}, 2016.

\bibitem{Materzynska2020SomethingElseCA}
Joanna Materzynska, Tete Xiao, Roei Herzig, Huijuan Xu, Xiaolong Wang, and Trevor Darrell.
\newblock Something-else: Compositional action recognition with spatial-temporal interaction networks.
\newblock In {\em CVPR}, pages 1046--1056, 2020.

\bibitem{Actor-Centric}
Effrosyni Mavroudi, Prashast Bindal, and René Vidal.
\newblock Actor-centric tubelets for real-time activity detection in extended videos.
\newblock In {\em WACVW}, pages 172--181, 2022.

\bibitem{Newell2017PixelsTG}
Alejandro Newell and Jia Deng.
\newblock Pixels to graphs by associative embedding.
\newblock In {\em NeurIPS}, 2017.

\bibitem{ou2022recognition}
Yangjun Ou, Li Mi, and Zhenzhong Chen.
\newblock Object-relation reasoning graph for action recognition.
\newblock In {\em CVPR}, 2022.

\bibitem{Rai2021HomeAG}
Nishant Rai, Haofeng Chen, Jingwei Ji, Rishi~M. Desai, Kazuki Kozuka, Shun Ishizaka, Ehsan Adeli, and Juan~Carlos Niebles.
\newblock Home action genome: Cooperative compositional action understanding.
\newblock In {\em CVPR}, 2021.

\bibitem{Ren2015FasterRT}
Shaoqing Ren, Kaiming He, Ross~B. Girshick, and Jian Sun.
\newblock Faster {R-CNN}: Towards real-time object detection with region proposal networks.
\newblock {\em IEEE TPAMI}, 39:1137--1149, 2015.

\bibitem{ryoo2021tokenlearner}
Michael Ryoo, AJ Piergiovanni, Anurag Arnab, Mostafa Dehghani, and Anelia Angelova.
\newblock Tokenlearner: Adaptive space-time tokenization for videos.
\newblock In M. Ranzato, A. Beygelzimer, Y. Dauphin, P.S. Liang, and J.~Wortman Vaughan, editors, {\em Advances in Neural Information Processing Systems}, volume~34, pages 12786--12797. Curran Associates, Inc., 2021.

\bibitem{assemblenetplusplus}
Michael~S. Ryoo, AJ Piergiovanni, Juhana Kangaspunta, and Anelia Angelova.
\newblock Assemblenet++: Assembling modality representations via attention connections.
\newblock In {\em ECCV}, 2020.

\bibitem{assemblenet}
Michael~S. Ryoo, AJ Piergiovanni, Tan Mingxing, and Anelia Angelova.
\newblock Assemblenet: Searching for multi-stream neural connectivity in video architectures.
\newblock In {\em ICLR}, 2020.

\bibitem{santoro2017simple}
Adam Santoro, David Raposo, David~G Barrett, Mateusz Malinowski, Razvan Pascanu, Peter Battaglia, and Timothy Lillicrap.
\newblock A simple neural network module for relational reasoning.
\newblock In {\em NeurIPS}, 2017.

\bibitem{shang2017video}
Xindi Shang, Tongwei Ren, Jingfan Guo, Hanwang Zhang, and Tat-Seng Chua.
\newblock Video visual relation detection.
\newblock In {\em ACM International Conference on Multimedia}, 2017.

\bibitem{sigurdsson2016hollywood}
Gunnar~A Sigurdsson, G{\"u}l Varol, Xiaolong Wang, Ali Farhadi, Ivan Laptev, and Abhinav Gupta.
\newblock Hollywood in homes: Crowdsourcing data collection for activity understanding.
\newblock In {\em ECCV}, 2016.

\bibitem{Tang2019LearningTC}
Kaihua Tang, Hanwang Zhang, Baoyuan Wu, Wenhan Luo, and W. Liu.
\newblock Learning to compose dynamic tree structures for visual contexts.
\newblock In {\em CVPR}, pages 6612--6621, 2019.

\bibitem{teng2021target}
Yao Teng, Limin Wang, Zhifeng Li, and Gangshan Wu.
\newblock Target adaptive context aggregation for video scene graph generation.
\newblock In {\em ICCV}, 2021.

\bibitem{TRACE}
Yao Teng, Limin Wang, Zhifeng Li, and Gangshan Wu.
\newblock Target adaptive context aggregation for video scene graph generation.
\newblock In {\em 2021 {IEEE/CVF} International Conference on Computer Vision, {ICCV} 2021, Montreal, QC, Canada, October 10-17, 2021}, pages 13668--13677. {IEEE}, 2021.

\bibitem{Wang2018NonlocalNN}
X. Wang, Ross~B. Girshick, Abhinav Gupta, and Kaiming He.
\newblock Non-local neural networks.
\newblock In {\em CVPR}, pages 7794--7803, 2018.

\bibitem{Wang2018VideosAS}
X. Wang and Abhinav Gupta.
\newblock Videos as space-time region graphs.
\newblock In {\em ECCV}, 2018.

\bibitem{Wu2019LongTermFB}
Chao-Yuan Wu, Christoph Feichtenhofer, Haoqi Fan, Kaiming He, Philipp Kr{\"a}henb{\"u}hl, and Ross~B. Girshick.
\newblock Long-term feature banks for detailed video understanding.
\newblock In {\em CVPR}, pages 284--293, 2019.

\bibitem{xietowards}
Wanze Xie, Junshen~K Chen, and Alan~Zelun Luo.
\newblock Towards compositional action recognition with spatio-temporal graph neural network.

\bibitem{IterativeMessagePassing@Danfei}
Danfei Xu, Yuke Zhu, Christopher~B. Choy, and Li Fei{-}Fei.
\newblock Scene graph generation by iterative message passing.
\newblock In {\em CVPR}, 2017.

\bibitem{GrapgRCNN@Jianwei}
Jianwei Yang, Jiasen Lu, Stefan Lee, Dhruv Batra, and Devi Parikh.
\newblock Graph {R-CNN} for scene graph generation.
\newblock In {\em ECCV}, 2018.

\bibitem{ZoomNet}
Guojun Yin, Lu Sheng, Bin Liu, Nenghai Yu, Xiaogang Wang, Jing Shao, and Chen~Change Loy.
\newblock Zoom-net: Mining deep feature interactions for visual relationship recognition.
\newblock In {\em Computer Vision - {ECCV} 2018 - 15th European Conference, Munich, Germany, September 8-14, 2018, Proceedings, Part {III}}, 2018.

\bibitem{zellers2018neural}
Rowan Zellers, Mark Yatskar, Sam Thomson, and Yejin Choi.
\newblock Neural motifs: Scene graph parsing with global context.
\newblock In {\em CVPR}, 2018.

\bibitem{zhang2019graphical}
Ji Zhang, Kevin~J. Shih, Ahmed Elgammal, Andrew Tao, and Bryan Catanzaro.
\newblock Graphical contrastive losses for scene graph parsing.
\newblock In {\em CVPR}, 2019.

\bibitem{ZhangLM21}
Yanyi Zhang, Xinyu Li, and Ivan Marsic.
\newblock Multi-label activity recognition using activity-specific features and activity correlations.
\newblock In {\em CVPR}, 2021.

\bibitem{zhu2018automatic}
Zhihao Zhu, Zhan Xue, and Zejian Yuan.
\newblock Automatic graphics program generation using attention-based hierarchical decoder.
\newblock In {\em Asian Conference on Computer Vision}, pages 181--196. Springer, 2018.

\end{thebibliography}
}

\end{document}